\documentclass[doublecol]{epl2vers2}
\usepackage{subfig}

\title{Self reference in word definitions}
\shorttitle{Self reference in word definitions} 

\author{David Levary\inst{1,3} \and Jean-Pierre Eckmann\inst{2} \and Elisha
Moses\inst{3} \and Tsvi Tlusty\inst{3}}
\shortauthor{D. Levary \etal}

\institute{
\inst{1}Department of Physics, Harvard University, 17 Oxford Street, Cambridge,
MA 02138\\
  \inst{2} D\'{e}partement de Physique Th\'{e}orique and Section de
Math\'{e}matiques, Universit\'{e} de Gen\`{e}ve, CH-1211, Geneva 4,
Switzerland\\
  \inst{3}Department of Physics of Complex Systems, Weizmann Institute of
Science, Rehovot 76100, Israel\\

}

\abstract{
Dictionaries are inherently circular in nature.  A given word is linked to a
set of alternative words (the definition) which in turn point to further
descendants. Iterating through definitions in this way, one typically finds
that definitions loop back upon themselves.  The graph formed by such
definitional relations is our object of study.  By eliminating those links
which are not in loops, we arrive at a core subgraph of highly connected nodes.
 We observe that definitional loops are conveniently classified by length, with
longer loops usually emerging from semantic misinterpretation.  By breaking the
long loops in the graph of the dictionary, we arrive at a set of disconnected
clusters.  We find that the words in these clusters constitute semantic units,
and moreover tend to have been introduced into the English language at similar
times, suggesting a possible mechanism for language evolution.}

\begin{document}

\maketitle

\section{Introduction}

Words are the building blocks of language.  By stringing together chains of
these simple blocks, complex thoughts and ideas can be conveyed.   For a
language to be effective, this transmission must not only be precise but also
efficient.  Indeed, the continuous expansion of human languages tends to be
driven less out of a need to express concepts that were previously
uncommunicable, than by the constraint that concepts be transmitted rapidly.
As a result of this need for efficient communication, the human lexicon is not
a simple 1 to 1 mapping of concepts onto words, but rather a complex web of
semantically related parts.

Network based formulations of human language have been employed previously to
study language evolution.  In this approach, words are considered to be the
nodes of a graph with edges drawn based on a variety of possible relationships
such as word co-occurrence in texts, thesauri, or word association experiments
on human users \cite{netreview,smallworld}.  Such language networks tend to be
scale-free and exhibit the small-world effect ({\it i.e.}, nodes are separated
from one another by a relatively small number of edges), characteristics shared
by many other complex, empirically observed networks \cite{netreview}.

 The notion of a dictionary based graph, in which directed links are drawn
between a word and the words in its definition, was proposed early on in view
of using computational tools \cite{Likowski}.  Dictionaries provide an
important tool for studying the relationship between words and concepts by
linking a given word to a set of alternative words (the definition) which can
express the same meaning.    Of course, the given definition is not unique.
One might just as well replace all of the words in the definition of the
original word in question, with their respective definitions. In the graph of
the dictionary then, a word and its set of descendants can be viewed as
semantically equivalent.

  Recently, the overall structure of this dictionary graph was analyzed
\cite{Hanard}.  It was found that dictionaries consist of a set of words,
roughly 10\% the size of the original dictionary, from which all other words
can be defined.  This subgraph was observed to be highly interconnected, with
a central strongly connected component dubbed the core.  The authors then
studied the connection of this finding with the acquisition of language in
children.

The existence of the core reflects an important property of the dictionary,
namely its requirement that every word have a definition ({\it i.e.}, a non-zero
out-degree).  The absence of ``axiomatic'' words whose definition is assumed
results in a graph with a large number of loops,  which is inherently not
tree-like in structure.  Here we study these definitional loops and show that
they arise not as simple artifacts of the dictionary's construction, but rather
as a manifestation of how coherent concepts are formed in a language.  The
distribution of loops in the actual dictionary differs markedly from the
predictions of random graph theory.  While the strong interconnectivity within
the core normally obscures semantic relationships among its elements, by
disconnecting the large loops of the graph, we are able to decompose the core
into semantically related components.  We show that by careful analysis of the
interactions among these components, some of the central  concepts upon which
vocabulary is structured can be revealed.  Finally, using additional
etymological data, we demonstrate that words within the same loop tend to have
been introduced into the English language at similar times, suggesting a
possible mechanism for language evolution.




\section{Dictionary Construction}

In order to construct an iterable dictionary, one must both reduce inflected
words to their stems and  resolve polysemous words to their proper sense.   We
therefore used as our primary dictionary eXtended WordNet, which provides
semantically parsed definitions for each WordNet 2.0 synset (set of synonymous
words) \cite{extended, wordnet}.  To reduce complexity, we chose to restrict
our attention to nouns as they are the part of speech generally most directly
related to the main concepts within a text \cite{Elisha}.

We treat the dictionary as a directed graph in which WordNet synsets are
designated as nodes, with a directed link drawn from a node to all of the
synset nodes which appear in its definition.  With this construction each sense
of a word is represented by a separate node.  The resulting graph consists of
79,689 nodes and 285,773 edges.  Its in-degree distribution obeys an
approximate power law, while the out-degree is distributed randomly following a
Poisson distribution.  The in-degree and out-degree distributions we observed
are consistent with those found in \cite{Hanard}.

For our studies, we found it convenient to represent this graph as an adjacency
matrix.  The process of iterating through definitions then corresponds to
taking successive powers of the adjacency matrix, with loops appearing as
non-zero entries in the diagonal.

\section{The Core}

\begin{figure}

\centering
\includegraphics[width=\columnwidth]{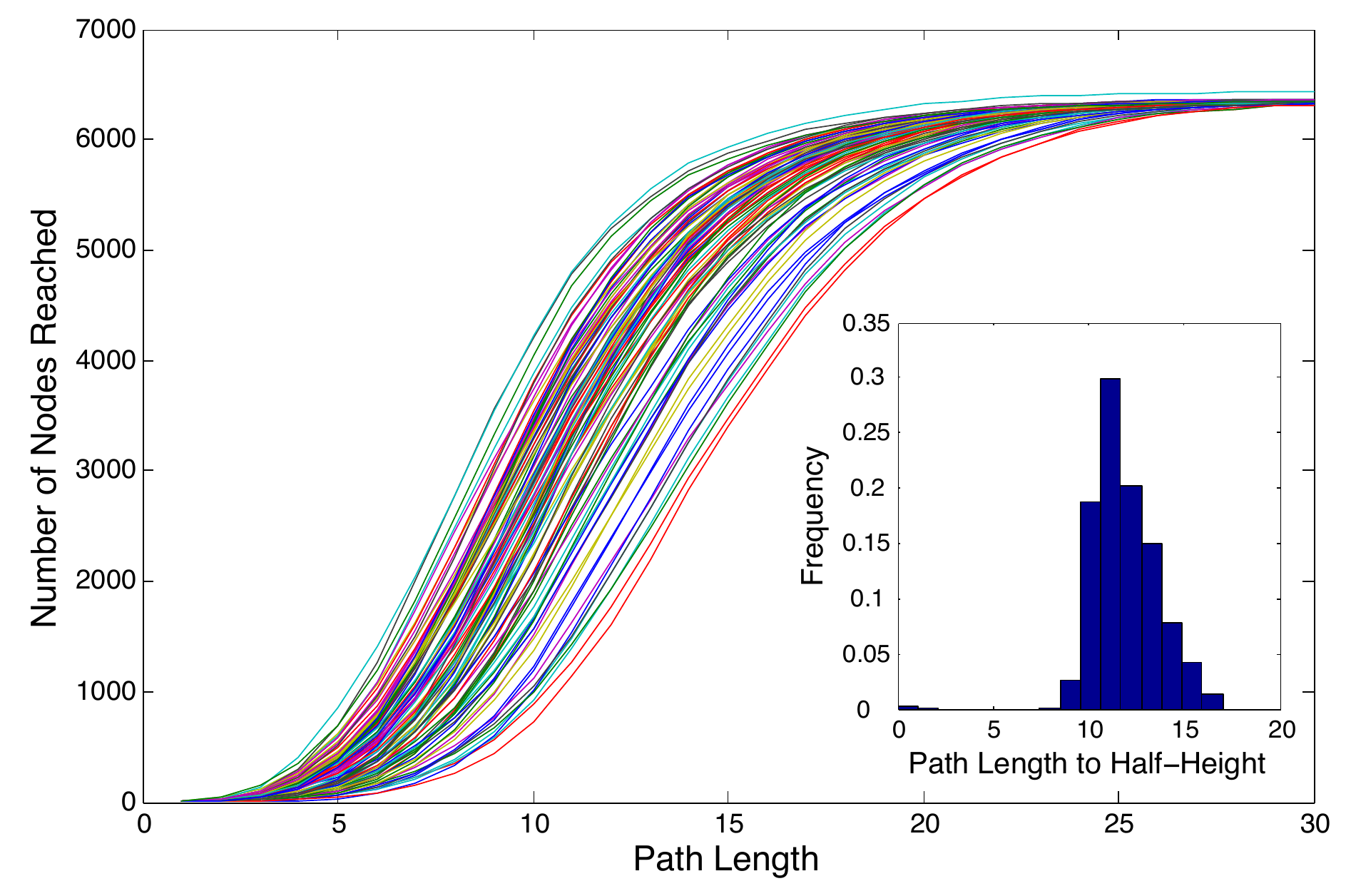}
\caption{Definitional iteration of words in the dictionary.  Using a random
sample of 100 words,  the number of unique nodes that could be reached within
the given distance of each node was recorded.  Nearly all words ultimately
reach a common set of 6,310 words which we call the core of the dictionary.
The convergence to the core is both rapid, as seen in the distribution of path
lengths to half height (inset), and complete,  indicating a very high
concentration of loops within the core.  Note that several words in our sample
were not connected to the core, existing instead as part of small isolated
definitional loops.  The half-height in these samples is therefore reached
almost immediately.  }
\label{iter}
\end{figure}

The current lexicon arose out of the need to express concepts both precisely
and concisely.  As such, there should exist not only words that expand the
breadth of ideas we are able to communicate, but also those that simply serve
to increase the efficiency of information transfer.  To isolate those words
which form the {\it conceptual} basis of the English language, we calculate the
``descendants'' of each word in the dictionary, namely all nodes which can
ultimately be reached along a directed path from the given starting point.
Surprisingly, as illustrated in fig.~\ref{iter}, we find that these sets of
descendants are almost completely independent of the starting point used to
reach them, intersecting in a 6,310 node set which we label as our
core\footnote{To ensure that this result was not an anomaly of WordNet glosses,
we constructed a graph using the English Wiktionary \cite{wiki} by associating
each word with the first sense of its definition.  Although this set is a mere
crude construction, a core of $\sim2,500$ words (out of $\sim80,000$) emerged
in an identical fashion.}.

While the existence of a central strongly connected has been demonstrated in
similar dictionary graphs previously \cite{Hanard}, the speed at which the
definitional paths converge on the core is surprising (inset to fig.
\ref{iter}).  After only twelve steps most paths will have already encompassed
half the core, and by thirty, all descendants will have already been reached.
This behavior suggests that the core contains a very high concentration of
overlapping loops because, otherwise, if there were disjoint
sinks, the algorithm would lead to one of them and miss the full height.

Our set of core words should theoretically be sufficient to define all words in
the dictionary, albeit with extensive paraphrasing, and thus can be thought of
as a simple vocabulary.  Having constructed this dictionary core by purely
computational means, it is interesting to compare the words in it to other
simple lexicons.  We compared our core to Basic English \cite{Ogden}, a set of
850 words British linguist Charles Ogden claimed sufficient for daily
discourse, as well as to the English translations of the words in J\={o}y\={o}
Kanji, the Japanese Education Ministry's list of 1,945 characters required to
be learned by Japanese secondary school students (accessed from \cite{Jap}).
As a control, we also compared these lists to the top 1000 most frequently used
words in all books found on Project Gutenberg (accessed from \cite{Gut}).  As
these lists were of course not sense disambiguated, we temporarily reduced the
resolution of our graph by making the nodes words (instead of synsets) and
using only the first sense of the definition.  Again, only nouns were
considered in all comparisons.

\begin{table}
\begin{center}
\scalebox{0.73}{
\begin{tabular}{|c|c|c|c|c|}
\hline
& {\bf Core} & {\bf Basic English} & {\bf J\={o}y\={o} Kanji} & {\bf
Gutenberg}\\
\hline\hline
 {\bf Core} & 1595 &314 (52\%) & 403 (29\%) & 265 (39\%)\\
{\bf Basic English} && 600 & 328 (24\%) & 213 (32\%)\\
 {\bf J\={o}y\={o} Kanji}& && 1376 & 319 (47\%)\\
 {\bf Gutenberg} & & && 673\\
\hline

\end{tabular}}
\caption{Intersection of core with other simple word lists. Table entries
represent the number of words in the intersection of the sets, with percent
overlap given in parentheses.  The core was reached using a simplified WordNet
dictionary graph, in which nodes were words (not synsets) with only the first
sense of the definition considered.   Only nouns in each word list were
considered.  Descriptions of the word lists are found in the main text. }
\end{center}
\label{simple}
\end{table}

Our new set of 1595 core words did share great overlap with all three lists
(see Table 1). Notably, however the overlap never exceeded 50\% of any word
 list.  A survey of the words in Basic English not found in the core reveals a
 trend of potentially useful but perhaps definitionally ``over-specific''
 words such as {\it apple, brick, chalk, hammer}, and {\it glove}.  While
 these words might come in handy in daily life, as Ogden had intended, it is
 easy to see how these words would be reduced in our dictionary  into more
 general words which in combination can communicate these more specific words
 ({\it e.g.}, in the case of {\it apple}, both {\it fruit} and {\it red}
 appear in our core).

\section{The Decomposition}

The manner in which we were able to reach the core suggests the somewhat
counterintuitive idea that all words are conceptually interconnected.  In order
to better characterize the connections which lead to the emergence of our core,
we searched for definitional loops within the dictionary.  We found that a
total of 9,085 nodes in our graph were elements of loops.  The core itself was
saturated with loops with over 99\% of its elements being involved in cycles
within it.

\begin{figure}

\centering
\includegraphics[width=\columnwidth]{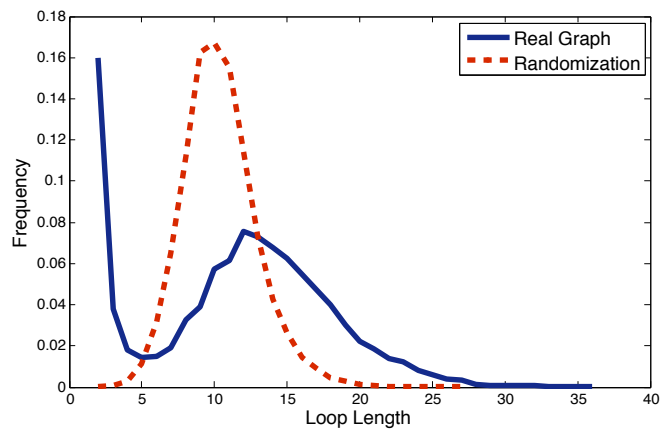}
\caption{Distribution of definitional loops in the dictionary.  The data
represent counts of links in the core indexed by the shortest loop in which
they appear.  For the randomization, links were randomly redrawn between nodes
while keeping the in-degree and out-degree distributions of the graph
constant.}
\label{loops}
\end{figure}

The distribution of loop lengths, shown in fig.~\ref{loops}, in the dictionary
is illuminating.  It appears that cycles in the dictionary fall into two
classes: short ($\leq 5$) and long ($>5$).  While the appearance of long loops
can be predicted solely based on the in and out degree distributions of our
graph (the randomization in the figure), the short loops appear to be a unique
feature arising from {\it meaningful} connections between nodes.  Inspection of
individual loops confirms this assessment, with small cycles following a very
clear conceptual path while large cycles are for the most part characterized by
one or more conceptual leaps, typically caused by a misinterpretation of word
sense as the following example illustrates.
\begin{center}
 $railcar \to rails \to bar \to weapon \to instrument \to skill \not\to train
\to railcar$
\end{center}
Though the link between bar and weapon is perhaps questionable, the link
between skill and train clearly is a case of mistaken sense, in this case
between ``train'' the verb and ``train'' the noun.  Such errors reflect the
fact that the semantic tagging in eXtended WordNet was done largely
computationally and is therefore subject to mistakes.  We have observed,
however, that the ratio of large to small loops is considerably lowered when links
are assigned based on the semantic tag in eXtended WordNet instead of being
assigned using na\"{i}ve, usage frequency based approaches (data not shown).

Fig.~\ref{loops} also shows a slight overabundance of links involved in large
loops in the dictionary as compared to the randomization.  This longer tail
appears to result from the fact that not all connections within a long loop are
false as  illustrated in the example loop above.  It therefore takes more
connections for a false loop to form in the real data than in the randomization
where {\it every} link is likely wrong.

\begin{figure*}[t]

\centering
\includegraphics[width=\textwidth, height=3in]{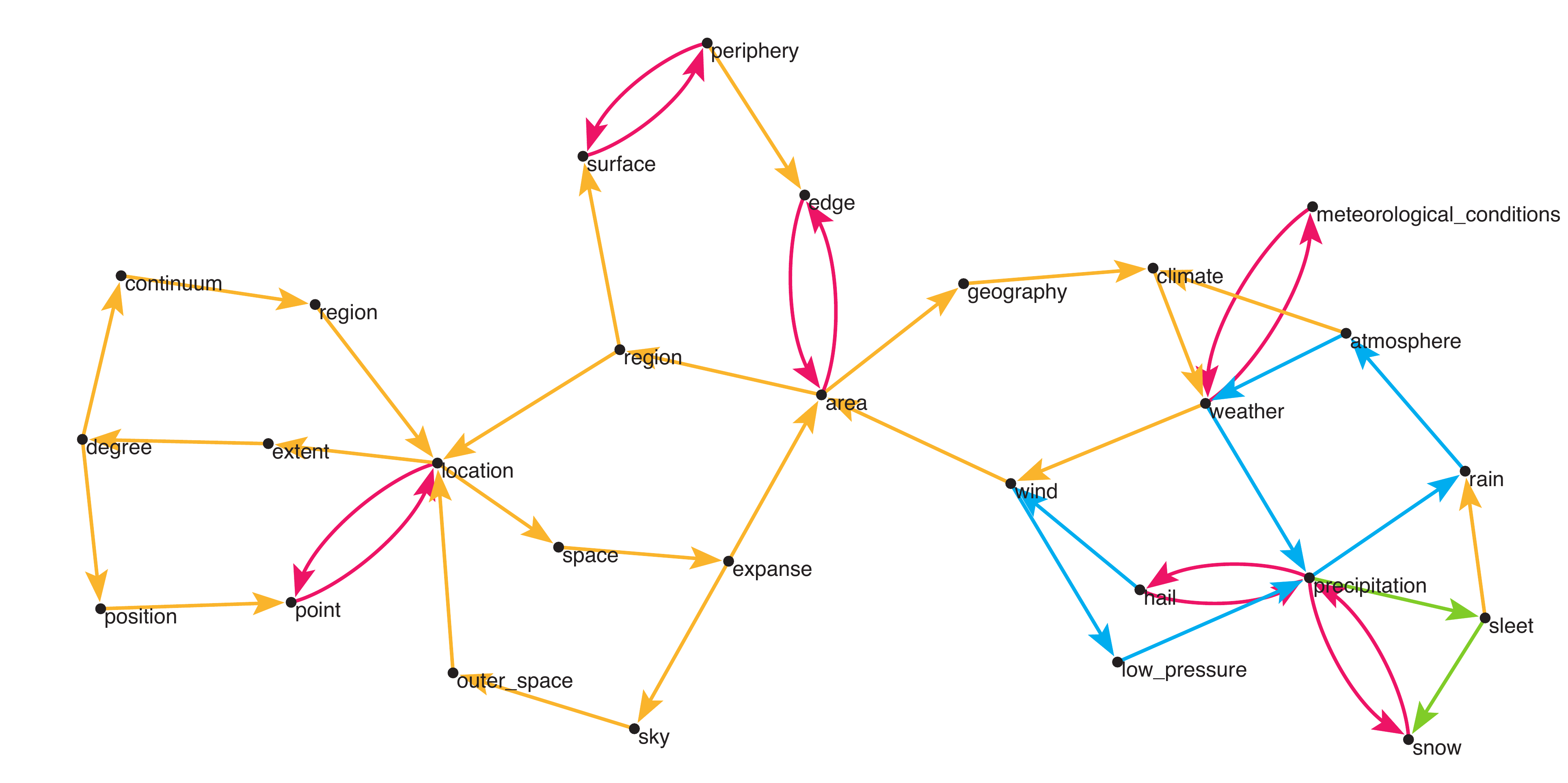}
\caption{An example of a large connected component in the decomposition. Arrows
are drawn from a node to words in it's definition.  Red links appear first in
two loops, green in three loops, blue in four loops, and orange in five loops.
}
\label{examplecomponent}
\end{figure*}

\begin{table}
\begin{center}
\scalebox{0.85}{
\begin{tabular}{|c|c|c|c|c|}
\hline \hline
emotion & height & bark & injury & winner\\
spirit & end & trunk & violence & contestant\\
dejection & dimension & tree & accident & competition\\
melancholy & length & lumber & & \\
feeling & &&&\\
\hline\hline
\end{tabular}}
\caption{Examples of strongly connected components in the decomposed core.}
\end{center}
\label{table:examples}
\end{table}

The finding that long loops generally emerge as a result of semantic
misinterpretations, suggests that the core is in fact structured among sets of
small, albeit interrelated, loops.  Indeed, when we considered only the links
in the core involved in small loops, we found that the core decomposed into
several hundred isolated, strongly connected components which show thematic
convergence (Table 2).  The size of these components ranged from 2 to 94. For
subsequent analysis we further resolved components with size greater than
twenty by considering only the links involved in four loops within them,
yielding a total of 386 components.

It is important to note that given the high-degree of connectivity between
loops, large loops did exist within some of these components.  The links within
these large loops, however, were simultaneously involved in small loops and as
a result generally followed a logical progression of ideas.
Fig.~\ref{examplecomponent} provides a graphical view of one of the
larger 
components in our decomposition, emphasizing the embedding of small loops
within the overall component.

Though clusters in our decomposition are built upon distinct semantic ideas,
they are {\it not} conceptually orthogonal to one another since different loops
actually share edges.  Meaningful connections between the connected components
do of course exist, our results suggesting simply that these connections are
generally acyclic in nature ({\it i.e.}, loops can be found only within
clusters, not between them).  In order to better characterize the interactions
among components and their role in the overall dictionary, we wish to
``define'' each word in the dictionary in terms of the semantic clusters.  To
quantify the importance of each component in the definition we count the number
of paths in our original graph leading from the word to a given cluster.  In an
attempt to increase the definitional weight of clusters located close to the
word in question, we allowed vertices and edges to be repeated when counting
paths so that the number of paths to a closer cluster continues to grow in the
time taken to reach a farther one.  This choice requires us to impose a bound
on the length of path we consider.  We choose this upper limit in path length
as 5, in keeping with our finding that loops of size greater than 5 usually
emerge from semantic misinterpretations.  Each node in the original graph can
now be associated with a vector whose elements are the number of paths from
that node to each of the 386 components. Concatenating these vectors yields a
sparse $79,689 \times 386$ matrix.

In analyzing this matrix we found that five components appeared in over 80\% of
the vectors.  Not surprisingly these components consisted of very general
words ({\it e.g.}, ``entity" and ``group") and were thus ignored in further
analysis and removed from the matrix.  In an attempt to identify cohesive
groups of connected components, we performed singular value decomposition
(SVD) on our matrix.  The resulting singular vectors (examples of which can
be found in Table 3) show a striking ability to capture major themes within
the dictionary including geography, life, and religion.  It is however the
connections between the elements in these singular vectors that are most
significant.  Though normally obscured by noisy connections in the
dictionary, links among topics such as the body, water, energy, and disease
in our singular vectors reflect powerful semantic chains underlying the {\it
conceptual} lexicon.

 \begin{table*}
\begin{center}
\scalebox{1}{
\begin{tabular}{c|c|c|c|c}
{\bf Vector 1} & {\bf Vector 2} & {\bf Vector 5} & {\bf Vector 6} & {\bf Vector
8}\\
\hline\hline
{\it Old World, oceans} & {\it spine, brain} & grains & {\it body parts} &
Jesus, Christianity\\
{\it bodies of water} & {\it body parts} & {\it bodies of water} & {\it flower,
seed} & man, woman\\
{\it The Americas} & Old World, oceans & herbage &  {\it tree, bark} & {\it
nucleus, DNA}\\
{\it influence, power} & {\it energy} & flower, seed & grains & Roman Empire\\
&pathology& {\it land} & nucleus, DNA & student, teacher\\
&{\it narrative} & {\it water} & spine, brain & speaker, speech\\
&{\it cognition} && {\it Gymnosperms} & Vatican, absolution\\
&{\it organic process} && bodies of water & Old Testament\\
&{\it respiration} && pathology & book\\
\end{tabular}}
\caption{Examples of the highest singular components for the dictionary.  The
elements in the singular components are the semantically cohesive clusters of
words obtained from decomposing the core.  For table entries, word(s)
representing the main theme of each cluster were chosen. Clusters are listed in
order of the absolute value of their coefficient in the singular component.
Only components whose absolute values were greater than 0.1 were listed.  Plain
text and italics indicate positive and negative values respectively.    The
3rd, 4th, and 7th highest singular components were very similar to the vectors
shown and therefore not displayed. }
\end{center}
\label{singular}
\end{table*}

\section{Loop Etymology}
As we have seen, definitional loops underlie much of the core structure of the
dictionary.  When one considers the evolution of a language, the question
arises how such loops in meaning came to exist.  Using the Online Etymology
Dictionary \cite{etymologydic}, we manually looked up the dates of origin for
words in small loops (namely the connected components in our decomposition).
Dates were recorded only when the definition given in the dictionary matched
the sense of the word in the loop and in the case of synsets with multiple
words, only the first word in the synset was used.  Given the considerable
vagueness surrounding dates of emergence in Old English, for the purposes of
our analysis all Old English words were recorded as having emerged in the year
1150.

\begin{figure}
\centering
\subfloat[]{\label{fig:pairwise}
\includegraphics[width=0.985\columnwidth]{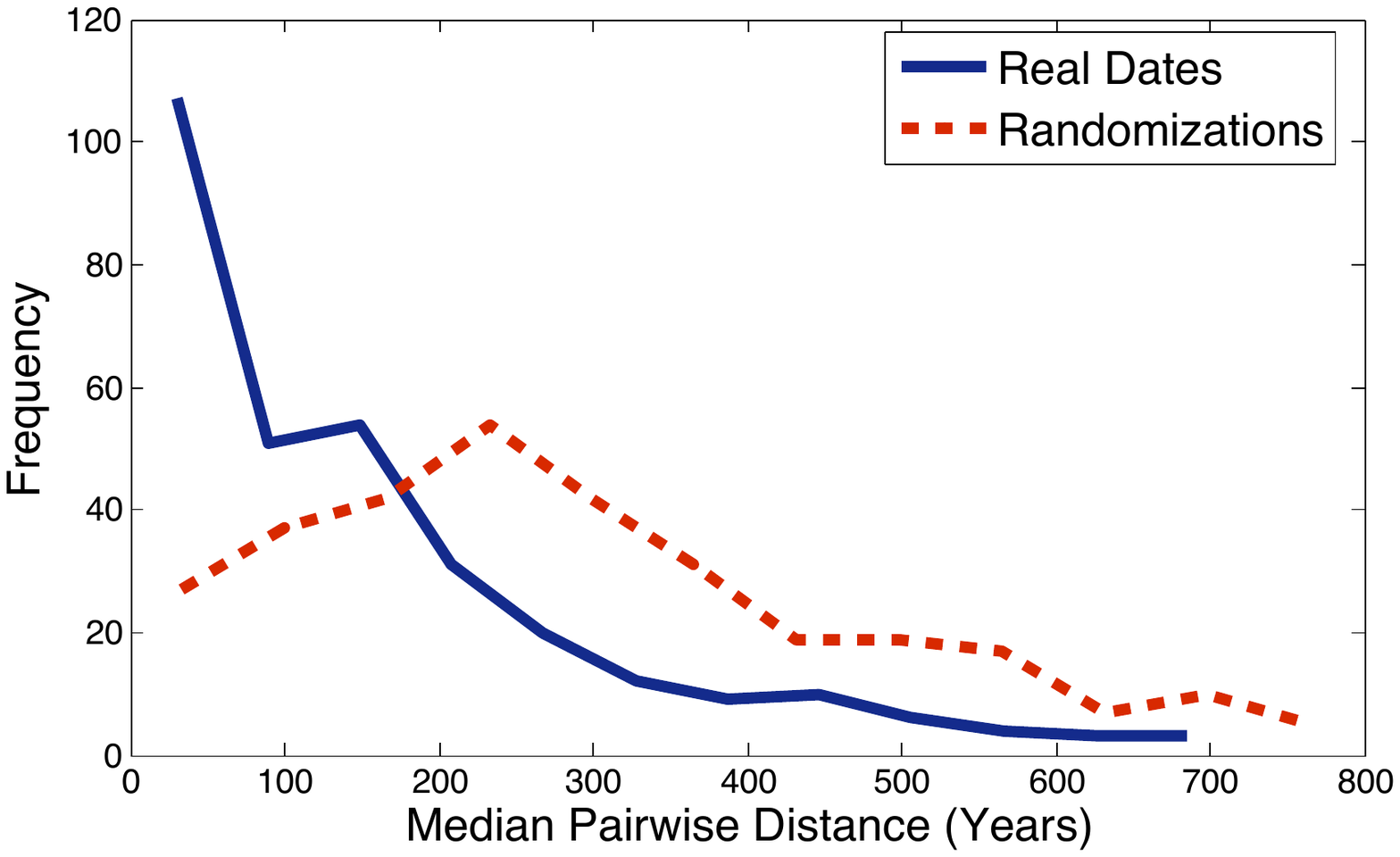}}\\
\subfloat[]{\label{fig:dates}\includegraphics[width=\columnwidth]{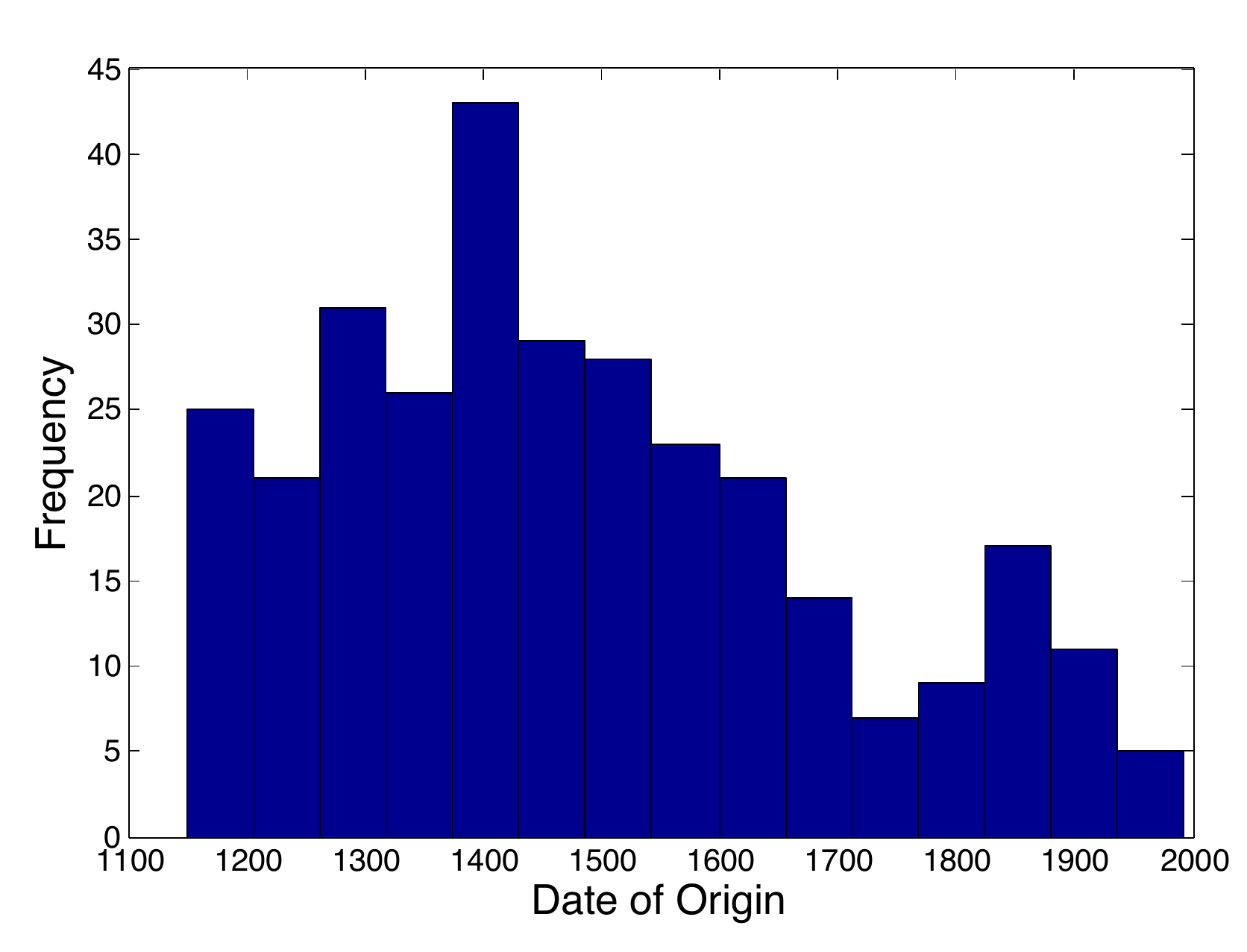}}
\caption{Dates of origin of words in loops. For each component in the
decomposition, the dates of origin of its element words (in the desired sense)
were looked up in the Etymology Dictionary.  Compound words and proper nouns
were ignored, as well as polysemous words. The median pairwise distance of
elements ({\it a}) and the mean date of origin ({\it b}) were calculated for
each of the 310 components in our analysis. }
\label{fig:test}
\end{figure}

After eliminating proper nouns and compound words, we found dates for 971 words
distributed among 310 connected components.  As shown in
fig.~\ref{fig:pairwise}, the distance among dates of origin of words in the
components is for the most part considerably smaller than that obtained by
randomly clustering these dates.  While several components do contain words
with somewhat disparate dates of origin, we found that such exceptions often
reflected fundamental changes in the understanding of a word after its
introduction.  For instance, the word ``cell'' was in use several hundred years
before the discovery of DNA, but since that event the two ideas have become
conceptually interdependent.  Interestingly, the distribution of mean dates of
origin for each component (fig.~\ref{fig:dates}) is bimodal in nature.  This
distribution is perhaps indicative of major periods of {\it conceptual}
expansion within the English language, with most growth appearing to occur
between the 14th-16th centuries, with a secondary growth of largely scientific
words emerging in the last two centuries.

The apparent coevolution of words in loops is quite striking.  While words in a
loop are of course semantically related, there is no a priori reason to assume
that semantically related words in general emerge around the same time period.
For instance, the word ``sneaker" is clearly closely related to the word
``shoe", yet it is not surprising that the two words emerged at very different
epochs (the Online Etymology Dictionary places sneaker in 1895 and shoe in Old
English).   The finding that words in loops are typically introduced into
language at the same time thus appears to reflect the unique type of semantic
relationship they share.

\section{Conclusions}

Dictionaries possess widespread circularity in definitions.  We have shown that
the loop structure of actual dictionaries varies dramatically from that which
would be predicted based on random graph theory alone.  Specifically, we found
that dictionaries rely on short loops of between two to five words in order to
define co-dependent concepts.  While long range loops do exist, they often
arise as a result of semantic misinterpretation.  Indeed, it appears to be
these false loops which account for the strongly connected nature of the
dictionary core, obscuring pockets of meaning within it.

In order to isolate ``true'' loops within the dictionary, we disconnected all
links between nodes which do not appear in loops of size five or smaller.  Due
to strong interconnectivity among certain loops, this approach importantly did
not lead to the dissolution of all long loops.  Rather, our graph decomposed
into a number of strongly connected components formed by collections of
overlapping, short-ranged loops which show thematic convergence.

Our finding that the words within a loop ({\it i.e.}, elements of the same
strongly connected component in our decomposition) were generally introduced
into the English language in the same time period underscores the unique
relationship among words involved in a definitional loop.  Although in theory
one need only know the meanings of some subset of the words in a loop in order
to infer the definitions of the remaining words,  at the conceptual level the
meanings of these words remain completely intertwined.

This of course begs the question of how loops could have come to exist in the
first place.  In order for a word to be introduced into language it must be
understood by multiple individuals to mean the same thing.  The necessary
synchronization of word meaning among different individuals is particularly
difficult when the meanings themselves exist as conceptual loops.  A potential
solution to this problem is for an individual to attempt to sequentially define
all the elements of the loop.  While the central concept of the loop cannot be
directly communicated, we propose that the juxtaposition of the partially
defined elements within the loop allows the receiver to infer the common link
among the words, thereby completing the definition of all words in the loop.
Such a system is consistent with our finding that words within a loop tend to
enter the lexicon at the same time and, if correct, suggests that definitional
loops are not simply a mathematical artifact of dictionaries, but rather a key
mechanism underlying language evolution.

\acknowledgments
We thank Ch.~Fellbaum for pointing us to the work of reference \cite{Hanard}
and for helpful discussions.  This work was funded in part by the Minerva
Foundation, Munich and the Einsten Minerva Center for Theoretical Physics.
JPE was partially supported by Fonds National Suisse and TT was supported
by Israel Science Foundation grant no. 1329/08.

\end{document}